\documentclass{article} 
\usepackage{nips15submit_e,times}
\usepackage{hyperref}
\usepackage{url}

\usepackage{graphicx}

\usepackage{tikz}
\usepackage{tkz-tab}
\usetikzlibrary{arrows,calc,shapes}

\usepackage{amsmath}
\usepackage{amssymb}

\usepackage{pgfplots}
\usepackage{pgfplotstable}

\usepackage{caption}
\usepackage{subcaption}

\title{Deep Attention Recurrent Q-Network}

\author{
Ivan Sorokin, Alexey Seleznev, Mikhail Pavlov, Aleksandr Fedorov, Anastasiia Ignateva \\
5vision\thanks{The authors are members of the 5vision team from the hackathon DeepHack.Game 2015.} \\
\texttt{5visionteam@gmail.com} \\
}

\nipsfinalcopy 

\begin{document}

\maketitle

\begin{abstract}
  A deep learning approach to reinforcement learning led to a general
  learner able to train on visual input to play a variety of arcade
  games at the human and superhuman levels. Its creators at the Google
  DeepMind's  team  called the  approach:  Deep Q-Network (DQN). We
  present an extension of DQN by ``soft'' and ``hard'' attention mechanisms.
  Tests of  the proposed  Deep Attention  Recurrent Q-Network (DARQN) algorithm on multiple Atari 2600 games show level
  of  performance  superior  to that of DQN. Moreover,  built-in
  attention  mechanisms  allow  a  direct  online  monitoring  of the
  training process by highlighting the  regions of the game screen the
  agent is focusing on when making decisions.
\end{abstract}

\section{Introduction and Related Work}

The recent success of Deep Q-Learning (DQL) in mastering human-level control policies on a variety of different Atari 2600 games \cite{mnih2015human} inspires artificial intellegence researchers to seek possible improvements to Google DeepMind's algorithm in order to further enhance its learning abilities \cite{hausknecht2015drqn, nair2015massively, Hasselt2015deep}. The goal of this concise paper is to present the authors' approach to addressing this challenge by providing DQN,
a deep neural network used in DQL as an analogue of a classic action-utility function, with such tools of modern machine learning as Long Short-Term Memory (LSTM) \cite{hochreiter1997LSTM} and visual attention mechanisms \cite{xu2015show, mnih2014ram, ba2015attention}.

Despite impressive results achieved by the Google DeepMind's intelligent agent, there are a number of elements to be improved in the existing 
algorithm. In particular, Hausknecht and Stone \cite{hausknecht2015drqn} pointed out that in practice, DQN decides on the next optimal 
action based on the visual information corresponding to the last four game states encountered by the agent. Therefore, the algorithm cannot 
master those games that require a player to remember events more distant than four screens in the past. It is for this reason that Hausknecht 
and Stone proposed the Deep Recurrent Q-Network (DRQN), a combination of LSTM and DQN in which (i) the fully connected layer in the latter is 
replaced for a LSTM one, and (ii) only the last visual frame at each timestep is used as DQN's input. The authors report that despite seeing 
only one visual frame, DRQN is still capable integrating relevant information across the frames. Nonetheless, no systematic improvement
in Atari game scores over the results of Mhih et al. \cite{mnih2015human} was observed.

Another drawback of DQN is its long training time, which is a critical component to the researchers' ability to carry out experiments 
with different network architectures and algorithm's parameter settings. According to \cite{mnih2015human}, it takes 12-14 days on a GPU 
to train the network. Nair et al. \cite{nair2015massively} proposed a new massively parallel version of the algorithm geared to address 
this problem. They report that its performance surpassed non-distributed DQN in 41 of the 49 games. However, extensive parallelization is 
not the only and, probably, not the most efficient remedy to the problem.

Recent achievements of visual attention models in caption generation \cite{xu2015show}, object tracking \cite{denil2012learning, mnih2014ram},
and machine translation \cite{bahdanau2014neural} have induced the authors of this paper to conduct a series of experiments so as to assess
possible benefits from incorporating attention mechanisms into the structure of the DRQN algorithm. The main advantage of utilizing these
mechanisms is that DRQN acquires the ability to select and then focus on relatively small informative regions of an input image, thus helping to
reduce the total number of parameters in the deep neural network and computational operations needed for training and testing it. In contrast to
DRQN, in this case, LSTM layer stores the data used not only for making decision on the next action, but also for choosing the next
region of attention. In addition to computational speedups, attention-based models can also add some degree of interpretability to the Deep Q-Learning process by providing
researchers with an opportunity to visualize ``where'' and ``what'' the agent's attention is focusing on.

The rest of the paper is organized as follows. In Section~\ref{sec:model}, two variants of the suggested DARQN algorithm are described. 
The results of applying the DARQN to two popular Atari 2600 games are presented in Section~\ref{sec:exp}. Conclusions are formulated in 
Section~\ref{sec:concl}.

\section{Deep Attention Recurrent Q-Network}
\label{sec:model}

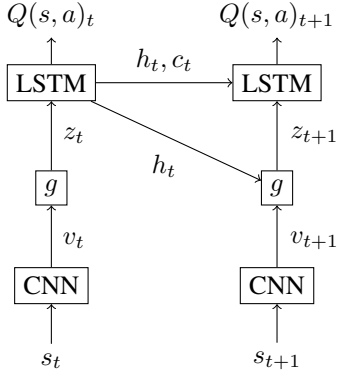
\begin{figure}
\centering
\begin{tikzpicture}
	\node(q1) at (0,4.6) {$Q(s,a)_t$};
	\node(q2) at (3,4.6) {$Q(s,a)_{t+1}$};
	\node[draw] (LSTM1) at (0,3.7) {LSTM};
	\node[draw] (LSTM2) at (3,3.7) {LSTM};
	\node at (0.3,3.0) {$z_t$};
	\node at (3.5,3.0) {$z_{t+1}$};
	\draw[->] (LSTM1) to (q1);
	\draw[->] (LSTM2) to (q2);
	\node at (1.5,4.0) {$h_t,c_t$};
	\draw[->] (LSTM1) to (LSTM2);
	\node[draw] (ATT1) at (0,2.3) {$g$};
	\node[draw] (ATT2) at (3,2.3) {$g$};
	\node at (1.5,2.6) {$h_t$};
	\draw[->] (LSTM1) to (ATT2);
	\draw[->] (ATT1) to (LSTM1);
	\draw[->] (ATT2) to (LSTM2);
	\node[draw] (CNN1) at (0,1) {CNN};
	\node[draw] (CNN2) at (3,1) {CNN};
	\node at (0.3,1.6) {$v_t$};
	\node at (3.5,1.6) {$v_{t+1}$};
	\draw[->] (CNN1) to (ATT1);
	\draw[->] (CNN2) to (ATT2);
	\node(s1) at (0,0) {$s_t$};
	\node(s2) at (3,0) {$s_{t+1}$};
	\draw[->] (s1) to (CNN1);
	\draw[->] (s2) to (CNN2);
\end{tikzpicture}
\caption{The Deep Attention Recurrent Q-Network}
\label{fig:model}
\end{figure}

The DARQN architecture is schematically shown in Figure~\ref{fig:model} and consists of three types of networks:
convolutional (CNN), attention, and recurrent. At each time step $t$, CNN receives a representation
of the current game state $s_t$ in the form of a visual frame, based on which it produces a set of $D$ feature maps, each having a dimension of $m\times{m}$. The attention network transformes these maps into a set of vectors $v_t = \{v_t^1,...,v_t^L\}, v_t^i\in{\mathbb{R}^D}, L=m*m$ and outputs
their linear combination $z_t\in{\mathbb{R}^D}$, called a context vector. The recurrent network, in our case LSTM,
takes as input the context vector, along with the previous hidden state $h_{t-1}$ and memory state $c_{t-1}$, and produces hidden state $h_t$ that is used by (i) a linear layer for evaluating $Q$-value of each action $a_t$ that the agent can take being in state $s_t$, (ii) the attention network for generating a context vector at the next time step $t+1$.
In the following subsections, we consider two approaches to the context vector calculation. As will be shown, they have important differences in the training procedure. \par

\subsection{Soft attention}

The ``soft'' attention mechanism assumes that the context vector $z_t$ can be represented as a weighted sum of all vectors $v_t^i$, 
each of which corresponds to the features extracted by CNN at different image regions. Weights in this sum are chosen in proportion to the vectors relative importance assessed by the attention network $g$. The $g$ network contains two fully connected layers followed by a softmax activation. 
Its output may be written as:
\begin{equation}
	g(v_t^i, h_{t-1}) = exp(Linear(Tanh(Linear(v_t^i) + W h_{t-1})))/Z,
	\label{eq:att}
\end{equation}
where $Z$ is a normalizing constant, $W$ is a weights matrix, $Linear(x) = A x + b$ is an affine transformation with some weights matrix $A$ and bias $b$. 
Once we have defined the importance of each location vector $v_t^i$, we can calculate the context vector $z_t$:
\begin{equation}
	z_t = \sum_{i=1}^L g(v_t^i, h_{t-1})v_t^i.
    \label{eq:soft}
\end{equation}
Other networks depicted in Figure~\ref{fig:model} have a standard form, the details of their realization are discussed in Section~\ref{sec:exp}. 
The whole DARQN model is trained by minimizing a sequence of loss functions:
\begin{equation}
	J_t(\theta_t) = \mathbb{E}_{s_t,a_t\sim\rho(\cdot),r_t}[(\mathbb{E}_{s_{t+1}\sim\mathcal{E}}[Y_t\mid{s_t,a_t}] - Q(s_t,a_t;\theta_t))^2],
	\label{eq:loss1}
\end{equation}
where $Y_t = r_t + \gamma \max_{a_{t+1}}Q(s_{t+1},a_{t+1};\theta_{t-1})$ is an approximate target value, $r_t$ is an immediate reward after taking action $a_t$ in state $s_t$, $\gamma \in [0,1]$ is a discount factor, $\mathcal{E}$ is an environment distribution, $\rho(s_t,a_t)$ is a behaviour distribution selected as $\epsilon$-greedy strategy, $\theta_t$ is a vector of all DARQN weights, including those belonging to the attention network. To optimize the loss function, we use the standard Q-learning update rule:
\begin{equation}
	\theta_{t+1} = \theta_t + \alpha(Y_t - Q(s_t,a_t;\theta_t))\nabla_{\theta_t}Q(s_t,a_t;\theta_t)
    \label{eq:req}
\end{equation}
All functions in DARQN are differentiable; therefore, the gradient exists for each parameter, and the whole model can be trained end-to-end.  The suggested algorithm also utilizes two training techniques proposed by Mnih et al. \cite{mnih2015human}, namely target network and experience replay. \par

\subsection{Hard attention}

The ``hard'' attention mechanism requires sampling only one attention location from $L$ available at each time step $t$ in accordance with some
stochastic attention policy $\pi_g$. 
In our case, this policy is represented by the neural network $g$ whose output \eqref{eq:att} consists of location selection probabilities and whose weights are the policy parameters. 
In order to train a network with stochastic units, the statistical gradient-following algorithm REINFORCE \cite{williams1992simple} may be used. In literature \cite{ba2015attention, xu2015show}, there are several successful examples 
of integrating this algorithm with Deep Learning. Unlike models proposed in these papers and trained by maximizing likelihood, the suggested algorithm is trained by minimizing a sequence of loss functions \eqref{eq:loss1}. Therefore, its training process 
is different. Assume that $s_t$ (and therefore $v_t$) was sampled from the environment distribution affected by 
the attention policy $\pi_g(i_t \mid v_t,h_{t-1})$, a categorical distribution with parameters given by a softmax 
layer \eqref{eq:att} of the attention network $g$. Then, in the policy gradient approach \cite{sutton2000policy}, updates 
of the policy parameters may be written as: 
\begin{equation}
	 \Delta \theta_t^g \propto \nabla_{\theta_t^g} \log \pi_g(i_t \mid v_t,h_{t-1}) R_t,
\end{equation}
where $R_t$ is a future discounted return after the agent selects the attention location $i_t$. In order to approximate this value, 
a separate neural network $G_t=Linear(h_t)$ has been introduced. This network is trained by regressing towards the expected value of $Y_t$. 
The final update rule for the attention network's parameters has the following form:
\begin{equation}
	\theta_{t+1}^g = \theta_t^g + \alpha \nabla_{\theta_t^g} \log \pi_g(i_t \mid v_t,h_{t-1}) (G_t - Y_t)
        \label{eq:prop}
\end{equation}
where the expression $G_t - Y_t$ can be interpreted in terms of \emph{advantage} function estimation \cite{schulman2015high}. Training \eqref{eq:prop} can also be described \cite{mnih2014ram} as adjusting the parameters $\theta_t^g$ of the attention network so that the log-probability of attention location $i_t$ that has led to a higher expected future reward is increased, while that of locations having produced a lower reward is decreased. In order to reduce a high variance of the stochastic gradient, a practical trick proposed in \cite{xu2015show} is utilized. At each time step, the context vector $z_t$ is found based on \eqref{eq:soft} with a 50\% chance. On the other hand, adding the entropy term on the categorical distribution has not resulted in any positive changes.

It is worth noting that for the hard attention DARQN model, CNN weights were preinitialized based on the corresponding weights of the trained soft attention model. In addition, the error backpropogation process does not affect weights at the previous time step, but does involve weights in convolutional layers. The latter receive the sum of two gradients: one from the attention network \eqref{eq:prop} and the other from the recurrent network \eqref{eq:req}.

\section{Experiments}
\label{sec:exp}

	\pgfplotstableread[col sep = comma]{breakout_result/dqn.txt}{\breakoutdqn}
	\pgfplotstableread[col sep = space]{breakout_result/drqn.txt}{\breakoutdrqn}
	\pgfplotstableread{breakout_result/darqn_soft.txt}{\breakoutsoft}
	\pgfplotstableread{breakout_result/darqn_soft10_lr01.txt}{\breakoutsoftten}
	\pgfplotstableread{breakout_result/darqn_hard.txt}{\breakouthard}
	\pgfplotstableread{breakout_result/darqn_hard8.txt}{\breakouthardeight}
	\pgfplotstableread{breakout_result/darqn_hard10.txt}{\breakouthardten}
	
	\pgfplotstableread{seaquest_result/dqn.txt}{\seaquestdqn}
	\pgfplotstableread{seaquest_result/drqn.txt}{\seaquestdrqn}
	\pgfplotstableread{seaquest_result/darqn_soft.txt}{\seaquestsoft}
	\pgfplotstableread{seaquest_result/darqn_hard.txt}{\seaquesthard}
	
	\pgfplotstableread{space_invaders_result/dqn.txt}{\spacedqn}
	\pgfplotstableread{space_invaders_result/drqn.txt}{\spacedrqn}
	\pgfplotstableread{space_invaders_result/darqn_soft.txt}{\spacesoft}
	\pgfplotstableread{space_invaders_result/darqn_hard.txt}{\spacehard}
	
	\pgfplotstableread{tutankham_result/dqn.txt}{\tutankhamdqn}
	\pgfplotstableread{tutankham_result/drqn.txt}{\tutankhamdrqn}
	\pgfplotstableread{tutankham_result/darqn_soft.txt}{\tutankhamsoft}
	\pgfplotstableread{tutankham_result/darqn_hard.txt}{\tutankhamhard}
	
	\pgfplotstableread{gopher_result/dqn.txt}{\gopherdqn}
	\pgfplotstableread{gopher_result/drqn.txt}{\gopherdrqn}
	\pgfplotstableread{gopher_result/darqn_soft.txt}{\gophersoft}
	\pgfplotstableread{gopher_result/darqn_hard.txt}{\gopherhard}

\begin{table}[t]
\caption{The best average reward per episode of 100 epochs for the four models on five Atari games. One epoch corresponds to $50,000$ steps. The hard and soft attention models as well as DRQN are trained with 4 unroll steps. DRQN weights are updated at each step, whereas DQN and DARQN weights are updated one time per 4 steps.}
\label{tbl:scores}
\begin{center}
\begin{tabular}{lccccc}
\multicolumn{1}{c}{}  
	&\multicolumn{1}{c}{\bf Breakout}  
	&\multicolumn{1}{c}{\bf Seaquest} 
	&\multicolumn{1}{c}{\bf S. Invaders}
	&\multicolumn{1}{c}{\bf Tutankham}
	&\multicolumn{1}{c}{\bf Gopher}
\\ \hline \\
DQN         & 241 & 1,284 & 916 & 197 & 1,976\\
DRQN        &  72 & 1,421 & 571 & 181 & 3,512\\
DARQN hard  &  20 & 3,005 & 558 & 128 & 2,510\\
DARQN soft  &  11 & 7,263 & 650 & 197 & 5,356\\
\end{tabular}
\end{center}
\end{table}

\begin{figure}

     \begin{subfigure}[b]{0.50\textwidth}
          \centering
          \resizebox{\linewidth}{!}{
		  
	  		\begin{tikzpicture}
	  		\begin{axis}[
	  				width=8cm, height=4.2cm,
	  				samples=100,
	  				legend pos = north west,
	  				table/x expr = \coordindex,
	  				every axis plot post/.append style={mark=none}
	  			]
	  			\addplot[color=black] table[y index = {1}] {\breakoutdqn};
	  			\addplot[color=brown] table[y index = {0}] {\breakoutdrqn};
	  			\addplot[color=blue] table[y index = {0}] {\breakoutsoft};
	  			\addplot[color=red] table[y index = {0}] {\breakouthard};
	  			\legend{DQN, DRQN, soft, hard}
	  		\end{axis}
	  		\end{tikzpicture}
		  
		  }
          \caption{Breakout}
          \label{fig:A}
     \end{subfigure}
     \begin{subfigure}[b]{0.50\textwidth}
          \centering
          \resizebox{\linewidth}{!}{
		  
	  		\begin{tikzpicture}
	  		\begin{axis}[
	  				width=8cm, height=4.2cm,
	  				samples=100,
	  				table/x expr = \coordindex,
	  				every axis plot post/.append style={mark=none}
	  			]
	  			\addplot[color=black] table[y index = {0}] {\seaquestdqn};
	  			\addplot[color=brown] table[y index = {0}] {\seaquestdrqn};
	  			\addplot[color=blue] table[y index = {0}] {\seaquestsoft};
	  			\addplot[color=red] table[y index = {0}] {\seaquesthard};
	  		\end{axis}
	  		\end{tikzpicture}
		  
		  }  
          \caption{Seaquest}
          \label{fig:B}
     \end{subfigure}
	 \caption{The average reward per episode for the four models on two Atari games as a function of the number of training epochs.}
	 \label{fig:scores}
\end{figure}
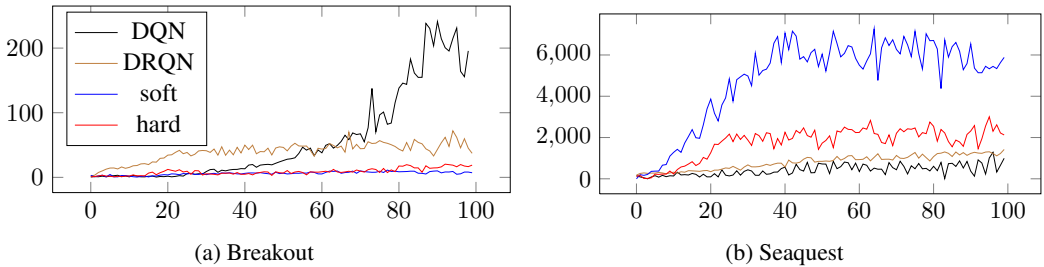

\begin{figure}
     \begin{subfigure}[b]{0.50\textwidth}
          \centering
          \resizebox{\linewidth}{!}{
	  		\begin{tikzpicture}
	  		\begin{axis}[
	  				width=8cm, height=4.1cm,
	  				samples=100,
	  				legend pos = north west,
	  				table/x expr = \coordindex,
	  				every axis plot post/.append style={mark=none}
	  			]
	  			\addplot[color=blue, dashed] table[y index = {0}] {\breakoutsoft};
	  			\addplot[color=blue] table[y index = {0}] {\breakoutsoftten};
	  			\legend{4 steps, 10 steps}
	  		\end{axis}
	  		\end{tikzpicture}
		  }
          \caption{Soft DARQN model}
          \label{fig:soft}
     \end{subfigure}
     \begin{subfigure}[b]{0.50\textwidth}
          \centering
          \resizebox{\linewidth}{!}{
	  		\begin{tikzpicture}
	  		\begin{axis}[
	  				width=8cm, height=4.1cm,
	  				samples=100,
	  				legend pos = north west,
	  				table/x expr = \coordindex,
	  				every axis plot post/.append style={mark=none}
	  			]
	  			\addplot[color=red, dashed] table[y index = {0}] {\breakouthard};
	  			\addplot[color=red] table[y index = {0}] {\breakouthardten};
				\legend{4 steps, 10 steps}
	  		\end{axis}
	  		\end{tikzpicture}
		  }  
          \caption{Hard DARQN model}
          \label{fig:hard}
     \end{subfigure}
	\caption{The average reward per episode for two DARQN models on Breakout with 4 and 10 unroll steps as a function of the number of training epochs. One epoch corresponds to $50,000$ steps. DARQN weights are updated one time per 4 steps. Weights of the CNN network in the hard attention model were preinitialized based on the weights of the trained soft attention model with 4 unroll steps.}
	\label{fig:unroll}
\end{figure}
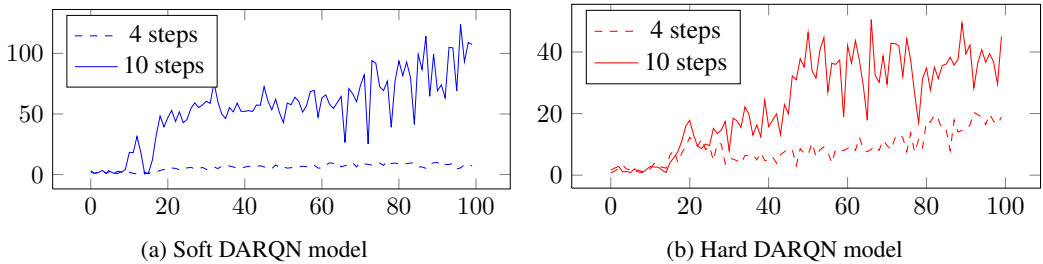

The proposed algorithm was tested on several popular Atari 2600 games: Breakout, Seaquest, Space Invaders, Tutankham, and Gopher. The results obtained were compared with the corresponding results of (i) DQN suggested by Mnih et al. \cite{mnih2015human} and implemented in Torch, (ii) DRQN suggested by Hausknecht and Stone \cite{hausknecht2015drqn} and implemented in Caffe. Our realization of DARQN is based on the source code \cite{mnih2015human} and is available online \footnote{\url{https://github.com/5vision/DARQN}}.

\subsection{Network Architecture}

The convolutional network architecture in DARQN is similar to that used in \cite{mnih2015human}, except for two peculiarities: 
its input is a $84\times84\times1$ tensor, and the output of its last (third) layer contains 256 feature maps $7\times7$. The attention network takes 49 vectors as input, each vector has a dimension of 256. The number of hidden units in the attention network is chosen to be equal to 256. The LSTM network also has 256 units, which is consistent with the number of attention network outputs.

It is intresting to compare the DARQN capacity to the capacities of DQN and DRQN. Depending on the game type, they may slightly differ. For Seaquest, a game with 18 possible actions, both DQN and DRQN (with 1 unroll step) have $1,693,362$ adjustable parameters, whereas the suggested hard and soft DARQN models have only $845,428$ and $845,171$ parameters, respectively.

\subsection{Hyper-parameters}

In all experiments, the discount factor was set to $\gamma = 0.99$, the learning rate $\alpha$ starts at $0.01$ and decays linearly to $0.00025$ over 1M steps for the soft attention model and from $0.001$ to $0.00025$ for the one with the hard attention model. The number of steps between target network updates was $10,000$. Training was done over 5M steps. The agent was evaluated after every $50,000$ steps based on the average reward per episode obtained by running an $\epsilon$-greedy policy with $\epsilon = 0.05$ for $25,000$ steps. The size of the experience replay memory was $500,000$ tuples. The memory was sampled to update the network every 4 steps with minibatches of size 32. The model was trained using the backpropogation through time. For each new minibatch, the initial LSTM's hidden and memory states were zeroed. To update weights $\theta_t$, the RMSProp algorithm with momentum equal to 0.95 was utilized. The simple exploration policy used was an $\epsilon$-greedy policy with the $\epsilon$ decreasing linearly from 1 to 0.1 over 1M steps.

\subsection{Results}

The main results of models comparison on the five Atari games are presented in Table~\ref{tbl:scores}. One can see that not on all of the games, the DARQN models achieve the results that are superior to corresponding results of DQN and DRQN. To provide some insight into advantages and disadvantages of the proposed models, the training process on the two games where DARQN obtains the best and the worst results is depicted in Figure~\ref{fig:scores}. 

On Seaquest, both DARQN models demonstrate a high level of performance. However, the hard attention-based agent seems to be inferior with respect to the soft one. In particular, it is unable to learn that in order to survive, the submarine has to regularly resurface. This problem can be attributed to one of the shortcomings of the policy gradient approach used in the hard attention mechanism's training procedure, namely to its tendency to converge to a local optimum. 

In the case of Breakout, models with LSTM have worse results than the original DQN. One possible reason for that is the low number of unroll steps used when training the LSTM network. To test this hypothesis, we repeated the whole experiment for the DARQN model with a greater number of unroll steps. The results presented in Figure~\ref{fig:unroll} show that despite some performance improvement, neither soft nor hard DARQN model can surpass the DQN results.

\begin{figure}
	\centering
	\includegraphics[width=2cm,keepaspectratio]{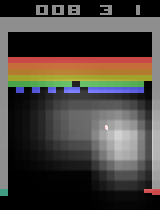}
	\includegraphics[width=2cm,keepaspectratio]{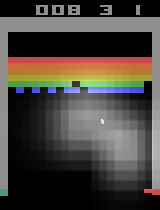}
	\includegraphics[width=2cm,keepaspectratio]{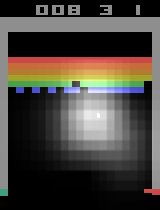}
	\includegraphics[width=2cm,keepaspectratio]{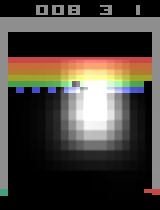}
	\includegraphics[width=2cm,keepaspectratio]{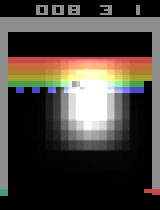}
	\includegraphics[width=2cm,keepaspectratio]{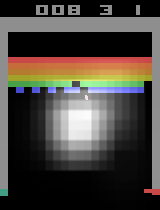}
	
	\includegraphics[width=2cm,keepaspectratio]{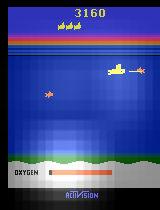}
	\includegraphics[width=2cm,keepaspectratio]{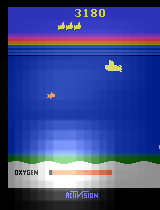}
	\includegraphics[width=2cm,keepaspectratio]{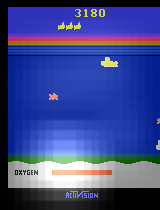}
	\includegraphics[width=2cm,keepaspectratio]{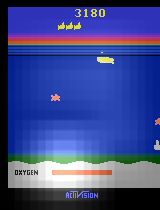}
	\includegraphics[width=2cm,keepaspectratio]{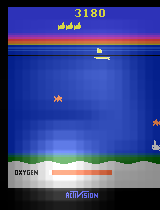}
	\includegraphics[width=2cm,keepaspectratio]{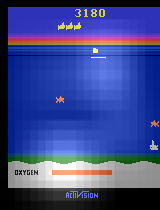}
	\caption{Visualization of attention regions for the soft DARQN model. \textbf{Top} row demonstrates the ability of the agent to focus on the ball trajectory in Breakout. \textbf{Bottom} row displays the process of submarine resurface in Seaquest. On the first screen, the agent mostly focuses on the oxygen indicator, but also notices enemies in its nearest vicinity. As the submarine rises to the surface, the attention of the agent switches to the submarine itself.}
	\label{fig:softframes}
\end{figure}
\begin{figure}
	\centering
	\includegraphics[width=2cm,keepaspectratio]{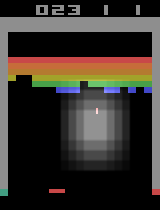}
	\includegraphics[width=2cm,keepaspectratio]{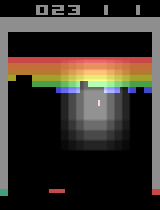}
	\includegraphics[width=2cm,keepaspectratio]{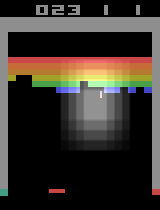}
	\includegraphics[width=2cm,keepaspectratio]{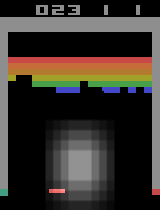}
	\includegraphics[width=2cm,keepaspectratio]{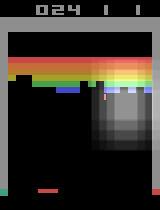}
	\includegraphics[width=2cm,keepaspectratio]{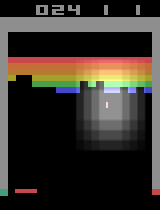}
	
	\includegraphics[width=2cm,keepaspectratio]{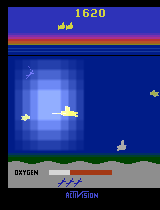}
	\includegraphics[width=2cm,keepaspectratio]{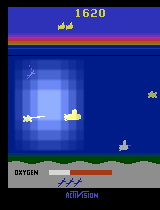}
	\includegraphics[width=2cm,keepaspectratio]{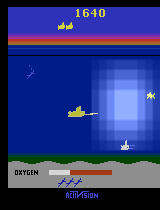}
	\includegraphics[width=2cm,keepaspectratio]{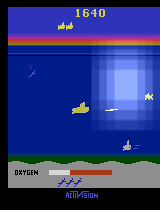}
	\includegraphics[width=2cm,keepaspectratio]{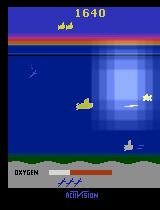}
	\includegraphics[width=2cm,keepaspectratio]{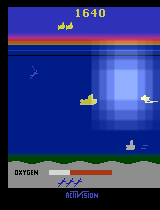}
	\caption{Visualization of attention regions for the hard DARQN model. \textbf{Top} row shows the agent's immediate response to the short-term ball disappearance in Breakout. \textbf{Bottom} row demonstrates the ability of the agent in Seaquest to focus attention on the enemy detected right up to the moment of its destruction.}
	\label{fig:hardframes}
\end{figure}

To visualize attention regions, we created $256$ subsidiary features maps $7\times7$ filled by output values \eqref{eq:att} and upsampled these maps through CNN layers, having the same structure as that used in DARQN model. The upsampled values were decreased to make an attention spot more transparent. In Figures~\ref{fig:softframes} and \ref{fig:hardframes}, some 
examples of highlighted attention regions are depicted. The corresponding game videos are available online \footnote{\url{https://www.youtube.com/playlist?list=PLKK-nv55ZMg583wK4Ny5sZu9YoFo27NBi}}.

\section{Conclusion and Future Work}
\label{sec:concl}

In this paper, we have presented one possible way of integrating attention mechanisms into the structure of Deep Q-Network. To test this model, a series of expirements was conducted on five Atari 2600 games. The results obtained allow us to arrive at conclusion that dispite having less optimized parameters, our model, at least on some Atari games, surpasses the results of the original DQN model, thereby demonstrating a greater generalization ability. Moreover, our attention-based algorithm allows gaining some insights into the logic of agent's behavior by displaying the regions of the game screen the agent is focusing on when making decisions.

Attention mechanisms can be considered as an additional filter gate in LSTM that processes structured visual data
produced by CNN for the entire image. Therefore, one promising direction of future research would be to apply multi-scale \cite{ba2015learning} or glimpse \cite{ba2015attention} visual attention mechanisms to DQN. The simple policy gradient-based algorithm, introduced for training the hard attention DARQN model, has shown a relatively poor level of performance. That is why another auspicious direction of future research would be (i) to test different techniques for reducing stochastic gradient variability \cite{schulman2015high}, (ii) to apply different approaches to training stochastic attention networks \cite{ba2015learning,schulman2015gradient}.

\subsubsection*{Acknowledgments}

We would like to thank Deep Knowledge Venture for financial support. In developing the ideas presented here, we have received helpful input from organizers of DeepHack.Game 2015 hackathon, especially from Sergey Plis (Datalytic Solutions). We also thank Greg Scantlen, CEO CreativeC.com, for letting us work on his private GPU cloud.

\bibliography{nips2015}{}
\bibliographystyle{unsrt}

\end{document}